\documentclass[10pt,twocolumn,letterpaper]{article}

\usepackage{wacv}
\usepackage{times}
\usepackage{epsfig}
\usepackage{graphicx}
\usepackage{amsmath}
\usepackage{amssymb}

\wacvfinalcopy 

\def\wacvPaperID{****} 

\ifwacvfinal\fi
\begin{document}

\title{Deep Adaptive Wavelet Network}

\author{Maria Ximena Bastidas Rodriguez \\
Universidad Nacional de Colombia\\
{\tt\small mxbastidasr@unal.edu.co}
\and
Adrien Gruson \\
The University of Tokyo and \\
McGill University\\
{\tt\small adrien.gruson@gmail.com}
\and
Luisa F. Polan\'ia\\
Target Corporation\\
{\tt\small lfpolani@udel.edu}
\and
Shin Fujieda\\
The University of Tokyo\\
{\tt\small sfujieda@graphics.ci.i.u-tokyo.ac.jp}
\and
Flavio Prieto Ortiz\\
Universidad Nacional de Colombia\\
{\tt\small faprietoo@unal.edu.co}
\and
Kohei Takayama\\
Digital Frontier Inc.\\
{\tt\small ktakayama@dfx.co.jp}
\and
Toshiya Hachisuka\\
The University of Tokyo\\
{\tt\small hachisuka@ci.i.u-tokyo.ac.jp}
}

\maketitle
\ifwacvfinal\thispagestyle{empty}\fi

\begin{abstract}
Even though convolutional neural networks have become the method of choice in many fields of computer vision, they still lack interpretability and are usually designed manually in a cumbersome trial-and-error process. This paper aims at overcoming those limitations by proposing a deep neural network, which is designed in a systematic fashion and is interpretable, by integrating multiresolution analysis at the core of the deep neural network design. By using the lifting scheme, it is possible to generate a wavelet representation and design a network capable of learning wavelet coefficients in an end-to-end form. Compared to state-of-the-art architectures, the proposed model requires less hyper-parameter tuning and achieves competitive accuracy in image classification tasks.
\end{abstract}

\section{Introduction}
\label{Section:Introduction}
%
%
Convolutional neural networks (CNNs) have become the dominant machine learning approach for image recognition. Numerous deep learning architectures have been developed ever since AlexNet~\cite{Alexnet} greatly outperformed other models on the ImageNet Challenge~\cite{ImageNet} in 2012. Based on backpropagation, CNNs can leverage correlation and structure inside datasets by directly tuning the network trainable parameters for a given task.

The trend in CNNs is to increase the number of layers to be able to model more complicated mathematical functions, to the point that recent architectures surpass 100 layers~\cite{ResNet,DenseNet}. There is, however, no guarantee that increasing the number of layers is always advantageous. Zagoruyko \etal~\cite{WideResNet} indeed showed that decreasing the number of layers and increasing the width of each layer leads to better performance than their commonly used thin and very deep counterpart, while reducing training time. Their results also support our general observation that current CNNs are not necessarily designed systematically, but usually through a manual process based on trial-and-error~\cite{Elsk17}.

A limitation of such networks is the lack of interpretability, which is usually referred to as the Achilles heel of CNNs. Convolutional neural networks are frequently treated as black-box function approximators which map a given input to a classification output~\cite{Dong17}. As deep learning becomes more ubiquitous in domains where transparency and reliability are priorities, such as healthcare, autonomous driving and finance, the need for interpretability becomes imperative~\cite{Chak17}. Interpretability enables users to understand the strengths and weaknesses of a model and conveys an understanding of how to diagnose and correct potential problems~\cite{Dong17}. Interpretable models are also considered less susceptible to adversarial attacks~\cite{Ross18}.

Theoretical properties of traditional signal processing approaches, such as multiresolution analysis using wavelets, are well studied, which makes such approaches more intepretable than CNNs. There are in fact several prior works that incorporate wavelet representations into CNNs. Oyallon \etal~\cite{Scattering} proposed a hybrid network which replaces the first layers of ResNet by a wavelet scattering network. This modified ResNet resulted in a comparable performance to that of the original ResNet but has a smaller number of trainable parameters. Williams \etal~\cite{Advanced} took the wavelet sub-bands of the input images as a new input and processed them with CNNs. In a different work~\cite{WaveletPooling}, they showed a wavelet pooling algorithm, which uses a second-level wavelet decomposition to subsample features. Lu \etal~\cite{DualTree} addressed the organ tissue segmentation problem by using a dual-tree wavelet transform on top of a CNN. Cotter and Kingsbury~\cite{WaveletDomain} also used a dual-tree wavelet transform to learn filters by taking activation layers into the wavelet space. 

Recently, Fujieda \etal~\cite{WCNN} proposed \emph{wavelet CNNs} (WCNNs), which were built upon the resemblance between multiresolution analysis and the convolutional filtering and pooling operations in CNNs. They proposed a CNN similar to DenseNet, but the Haar wavelets (which are commonly used in multiresolution analysis) were used as convolution and pooling layers. These wavelet layers were concatenated with the feature maps produced by the succeeding convolutional blocks. This model  is more interpretable than CNNs since the wavelet layers generate the wavelet transform of the input. The use of a fixed wavelet (Haar), however, is likely suboptimal as it restricts the adaptability and cannot leverage data-driven learning.

Inspired by WCNNs, we propose to perform multiresolution analysis within the network architecture by using the lifting scheme~\cite{Lifting1} to perform a data-driven wavelet transform. The lifting scheme offers many advantages compared to the first-generation wavelets, such as adaptivity, data-drivenness, non-linearity, faster and easier implementation, fully in-place calculation, and reversible integer-to-integer transform~\cite{Zhan06}. 

Unlike previous works which combine CNNs and wavelets, our model learns all the filters from data in an end-to-end framework. Due to the connection with multiresolution analysis, the number of layers in our network is determined mathematically.
The combination of end-to-end training and multiresolution analysis via the lifting scheme allows us to efficiently capture the essential information from the input for image classification such as texture and object recognition. The use of multiresolution analysis generates a relevant visual representation at each decomposition level, which contributes to the interpretability of the network.

The evaluation of the proposed network was performed on three competitive benchmarks for texture and object classification tasks, namely, \textit{KTH-TIPS-b}, \textit{CIFAR-10} and \textit{CIFAR-100}. The proposed model attains comparable results to those presented by the state-of-the-art on texture classification, trained end-to-end from scratch, with a fraction of the number of trainable parameters. Moreover, the proposed model shows better generalization compared to networks especially tailored for texture recognition as it presents good performance for object classification task. 
This work is the first to propose trainable wavelet filters in the context of CNNs. In summary, we propose a deep neural network for image classification which exhibits the following properties: 

The network is interpretable since approximation and detail coefficients, which have a relevant visual representation, are generated by the multiresolution analysis using the lifting scheme at each decomposition level.

The network extracts features using a multiresolution analysis approach and capture essential information for classification task reducing the number of trainable parameters in texture classification. The loss function used to train the network ensures that the captured information is relevant to the classification task. 

The architecture offers competitive accuracy in texture and object classification tasks.

\section{Background}
This section briefly describes multiresolution analysis and the lifting scheme which are the building blocks of our model.

\subsection{CNNs as Multiresolution Analysis}
\label{Section:CNN-MR}
Convolutional neural networks proposed by LeCun in 1989~\cite{LeCun} contain filtering and downsampling steps. In order to have a better understanding of CNNs, we propose to interpret convolution and pooling operations in CNNs as operations in multiresolution analysis~\cite{Mallat}. In the following, only one-dimensional input signals are considered for simplicity, but the analysis can be easily extended to higher dimensional signals.

Given an input vector $x=(x[0],x[1],...,x[N-1]) \in \mathbb{R}^N$, and a weighting function $\omega$, referred to as kernel, the convolution layer output (or feature map) $y=(y[0],y[1],...,y[N-1]) \in \mathbb{R}^N$ can be defined as
\begin{equation}\label{Equation:convolution}
    y[n]=(x*\omega)[n]=\sum_{j \in K}x[n+j]\omega[j] \,
\end{equation}
where $K$ is the set of kernel indices.

The role of the pooling layers is to output a summary statistic of the input~\cite{Goodfellow}. It is normally used to reduce the complexity and to simplify information. Most common pooling layers consist of convolution and downsampling in signal processing.
%
%
%
%
%
Using the standard downsampling symbol $\downarrow$, the output vector $o$ from a pooling layer can be written as
\begin{equation}\label{Equation:downsampling}
    o=(b*\mathbf{p})\downarrow p,
\end{equation}
where $\mathbf{p}=(1/p,...1/p)\in \mathbb{R}^p$ is the pooling filter.

We can now interpret convolution and pooling layers as operations in multiresolution analysis. In this analysis, the resolution of a signal (measure of the amount of detail in a signal) is changed by a filtering operation, and the scale of a signal is changed by a downsampling operation~\cite{Wavelet_Tutorial}. The wavelet transform, for example, repeatedly decomposes a signal into spectrum sub-bands by using low-pass $k_{l}$ and high-pass $k_{h}$ filters and applies downsampling by a factor of 2.

Then, to perform a multiresolution analysis, a new signal decomposition is obtained by taking as input the low-pass filtered sub-band $c_l$. Each of these decompositions are referred to as levels, and generate a hierarchical decomposition of the signal into $c_{l,t}$ and $d_{h,t}$ each time. Let $k_{l,t}$ and $k_{h,t}$ denote the low-pass and high-pass filters at step $t$, respectively. Such transformation is thus represented as a sequence of convolution and pooling operations,
\begin{equation}\label{Equation:Multiresolution}
\begin{split}
    c_{l,t+1}=(c_{l,t}*k_{l,t})\downarrow 2 \\
    d_{h,t+1}=(d_{h,t}*k_{h,t})\downarrow 2,
\end{split}
\end{equation}
where $c_{l,t+1}$ and $d_{h,t+1}$ denote the approximation and detail coefficients generated at step $t$, respectively, $c_{l,0}=x$ and $d_{h,0} = x$.
Based on this level decomposition-based construction, it is possible to compare CNNs structures with multiresolution analysis, as Eqns.~\ref{Equation:downsampling} and  \ref{Equation:Multiresolution} are quite similar, with the difference that in CNNs the filters are randomly selected and their output does not have a meaningful interpretation.
%
%
\begin{figure*}[t]
    \centering
    \includegraphics[width=0.7\linewidth]{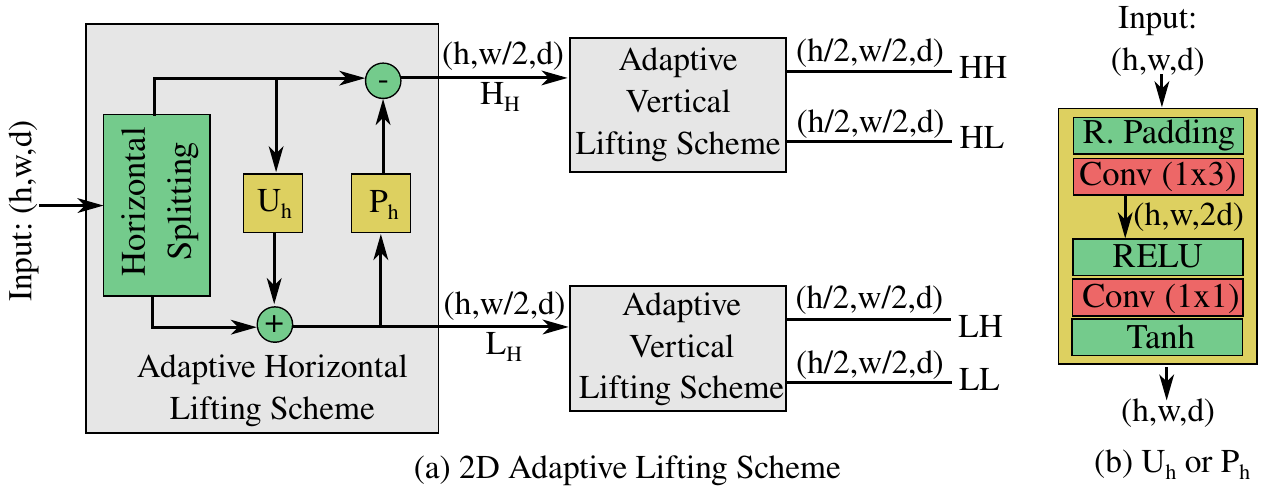}
    \caption{(a) The 2D Adaptive Lifting Scheme consists of successively applying horizontal and vertical lifting steps where each of them have their own predictor and updater. (b) The predictors and updaters are based on operations, such as paddings, convolutions, and non-linear activation functions, which can be either trainable (red boxes) or fixed (green boxes).}
    \label{fig:lifting_wavelet_pooling}
\end{figure*}
\subsection{Lifting Scheme}
\label{Section:Lifting_scheme}
The first-generation wavelets are mathematical functions that allow for efficient representations of data using only a small set of coefficients by exploiting space and frequency correlation~\cite{Wavelet_Tutorial}. The main idea behind the wavelet transform is to build a sparse approximation of natural signals through the correlation  structure present on them. This correlation is normally local in space and frequency, meaning that there is a stronger correlation among the neighboring samples on the signal. 
The construction of mother wavelets is traditionally performed by using the Fourier transform, however, this can also be constructed in the spatial domain~\cite{Lifting2}.

The lifting scheme, which is also known as second-generation wavelets~\cite{Lifting1}, is a simple and powerful approach to define wavelets that has the same properties as the first-generation wavelets~\cite{Lifting2}. 
The lifting scheme takes as input a signal $x$ and generates as outputs the approximation $c$ and the details $d$ sub-bands of the wavelet transform. Designing such lifting scheme consists of three stages~\cite{Lifting3} as follows.
%
%
\paragraph*{Splitting the signal.} This step consists of splitting the input signal into two non-overlapping partitions. The simplest possible partition is chosen; \textit{i.e.} the input signal $x$ is divided into even and odd components denoted as $x_e$ and $x_o$, respectively, and defined as $x_e[n]=x[2n]$ and $x_o[n]=x[2n+1]$.
%
%
\paragraph*{Updater.} This stage will take care of the separation in the frequency domain, looking that the approximation has the same running average as the input signal~\cite{Lifting2}. To achieve this, the approximation $c$ should be a function of the even part $x_e$ of the signal plus an update operator $U$.

Let $x_o^{L_U}[n]= x_o[n-L_U], x_o[n-L_U+1], \ldots, x_o[n+L_U-1], x_o[n+L_U]$ denote the sequence of $2L_U+1$ neighboring odd polyphase samples of $x_e[n]$. The even polyphase samples are updated using $x_o^{L_u}[n]$, and the result forms the approximation $c$, as described in Eqn.~\ref{Equation:Lifting_c}, where $U(\cdot)$ is the update operator.
\begin{equation} \label{Equation:Lifting_c}
c[n]=x_e[n]+U(x_o^{L_U}[n]).
\end{equation}
\paragraph*{Predictor.}The splitting partitions of the signals are, typically, closely correlated. Thus, given one of them, it is possible to build a good predictor $P$ for the other set, by tracking the difference (or details) $d$ among them~\cite{Lifting2}. As the even part of the signal $x[n]$ corresponds to the approximation $c[n]$ (Eqn.~\ref{Equation:Lifting_c}), then it is possible to define P as a function of $c[n]$.

Let $c^{L_P}[n]= c[n-L_P], c[n-L_P+1], \ldots, c[n+L_P-1], c[n+L_P]$ denote a sequence of $2L_P+1$ approximation coefficients. In the prediction step, the odd polyphase samples are predicted from $c^{L_P}[n]$. The resulting prediction residuals, or high sub-band coefficients $d$, are computed by Eqn.~\ref{Equation:Lifting_details}, where $P(\cdot)$ is the prediction operator.
\begin{equation} \label{Equation:Lifting_details}
d[n]=x_o[n]-P(c^{L_P}[n]).
\end{equation}
\subsubsection{Lifting Scheme Via Neural Networks}

Yi \etal~\cite{Adaptive_Lifting} proposed to replace the updater and the predictor with non-linear functions represented by neural networks to adapt to the input signals. To train them, the authors proposed to use the following loss functions:
\begin{equation}
\label{losses}
\begin{split}
     \mathbf{Loss(P)} &= \sum_n(P(c^{L_p}[n])-x_o[n])^2\\
     \mathbf{Loss(U)} &= \sum_n(U(x_o^{L_U}[n])-(x_o[n]-x_e[n]))^2,
\end{split}
\end{equation}
where $\mathbf{Loss(P)}$ and $\mathbf{Loss(U)}$ are the loss functions for the predictor and updater, respectively. The loss for the predictor network promotes the minimization of the detail coefficients magnitude (Eqn.~\ref{Equation:Lifting_details}). Yi \etal~\cite{Adaptive_Lifting} argued that $c$ is close to $x_e$ by definition, which only makes it necessary for the loss function of the updater network to minimize the distance between $c$ and $x_o$. Note that in Yi \etal~\cite{Adaptive_Lifting}, the predictor and the updater were trained sequentially.
\section{Deep Adaptive Wavelet Network (DAWN)} \label{Section:DAWN}
We propose a new network architecture, Deep Adaptive Wavelet Network~(DAWN), which uses the lifting scheme to capture essential information from the input data for image classification. The adaptive lifting scheme presented by Yi \etal~\cite{Adaptive_Lifting} showed that neural networks trained through backpropagation can be used to implement the lifting scheme for one-dimensional (1D) signals. The DAWN architecture extends this idea to address a classification task, and integrates multiresolution analysis into neural networks.
The proposed model performs multiresolution analysis at the core of the classification network by training the parameters of two-dimensional (2D) lifting schemes in an end-to-end fashion.
None of the previous wavelet-based CNN approaches have performed this end-to-end training while learning the wavelet parameters.
\subsection{2D Adaptive Lifting Scheme}
We first explain the proposed \emph{2D Adaptive Lifting Scheme}, and then present the integration of the 2D lifting scheme into the proposed classification architecture.

The \emph{2D Adaptive Lifting Scheme} consists of a horizontal lifting step followed by two independent vertical lifting steps that generate the four sub-bands of the wavelet transform. These sub-bands are denoted as LL, LH, HL, and HH, where L and H represent low and high frequency information, respectively, and the first and second positions refer to the horizontal and the vertical directions, respectively. Note that the 2D lifting scheme, illustrated in Figure~\ref{fig:lifting_wavelet_pooling}~(a), performs spatial pooling, as the spatial size of the outputs are reduced by half with respect to the input.

The \emph{Adaptive Horizontal Lifting Scheme} performs horizontal analysis by splitting the 2D signal into two non-overlapping partitions. We chose to partition the 2D signal into the even ($x_e[n]=x[2n]$) and odd ($x_o[n]=x[2n+1]$) horizontal components. Then a horizontal updater ($U_h$) and a horizontal predictor ($P_h$) operators are applied in the same way as described in Section \ref{Section:Lifting_scheme}. The vertical lifting step has a similar structure as the horizontal lifting step, but in this case, the splitting is performed in the vertical component of the 2D signal, followed by the processing, performed by the vertical updater $U_v$ and the vertical predictor $P_v$ operators.
%
%
\paragraph*{Predictor and Updater.}
The internal structure of the updater and the predictor is the same for both the vertical and horizontal directions. Figure~\ref{fig:lifting_wavelet_pooling}~(b) shows the structure of the horizontal predictor (or horizontal updater). At the beginning, reflection padding is applied instead of zero padding to prevent harmful border effects caused by the convolution operation. Then, a 2D convolutional layer, where the kernel size, depending on the direction of analysis ($(1,3)$ if horizontal while $(3,1)$ if vertical), is applied. The output depth of the first convolutional layer is set to be twice the number of channels of the input. Then, a second convolutional layer with kernels of size (1,1) is applied. The output depth of this layer is set the same as the initial input depth of the predictor/updater. The stride for all the convolutions is set to $(1,1)$. 
The first convolutional layer is followed by a $relu$ activation function, and we can benefit from its properties of sparsity and a reduced likelihood of vanishing gradient. The last convolutional layer is followed by a $tanh$ activation function as we do not want to discard negative values in this stage.
%

\paragraph*{Design Choices.} We arbitrarily chose to perform the horizontal analysis before the vertical analysis. However, there are no performance variations by computing the vertical analysis first.
The number of convolutional layers and the kernel size used in the predictor/updater will be discussed during the hyperparameter study (Section~\ref{section:hyper}).
The main concern while choosing the depth was to maintain a relevant visual representation of the approximation and details sub-bands, while not considerably increasing the number of network parameters.

\subsection{DAWN Architecture}
\begin{figure}[t]
    \centering
    \includegraphics[width=0.5\textwidth]{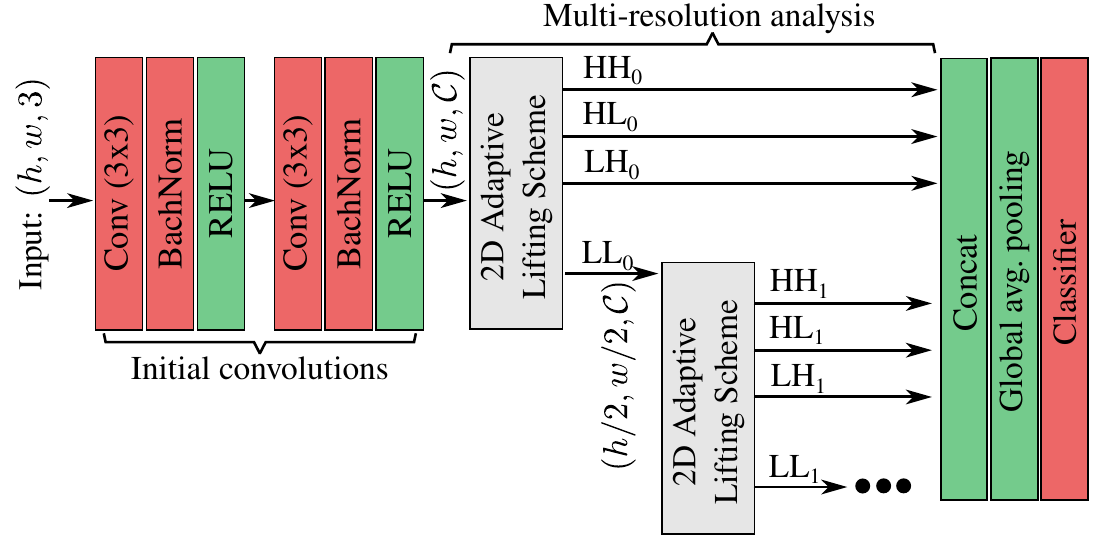}
    \caption{The proposed architecture is composed by three modules: \textit{i)} Initial convolutional layers to increase the input depth, \textit{ii)} $M$ levels of multiresolution analysis, where 2D lifting scheme is applied on the approximation output of the previous level, and \textit{iii)} a large concat of details from the different levels and the approximation, followed by a global average pooling and a dense layer. The operations in the architecture can be classified as either trainable (red boxes) or fixed (green boxes).}
    \label{fig:DAWN_architecture}
\end{figure}

The DAWN architecture is based on stacking multiple \emph{2D Adaptive Lifting Schemes} to perform multiresolution analysis (see Figure~\ref{fig:DAWN_architecture}). The architecture starts with two convolutional layers followed by a multiresolution analysis of $M$ levels. Each level consists of a \emph{2D Adaptive Lifting Scheme}, which generates as output the four wavelet transform sub-bands LL, LH, HL and HH, and the input correspond to the low level sub-band (LL) from the previous level. The details sub-bands from each level (LH, HL, HH) are concatenated and followed by a global average pooling layer~\cite{GAP}, used to reduce overfitting and to perform dimensionality reduction. In the last level, the global average pooling of the outputs at each level are concatenated before the final fully-connected layer and a log-softmax to perform the classification task.
%
%

\paragraph*{Number of levels.}
The minimum size of feature maps at the end of the network for this architecture is set to $4\times4$ as it is the minimum possible size that still maintains the 2D signal structure.
Assuming that the input images are square, the number of levels $M$, is given by $M=\left \lfloor{\mathrm{log}_2(i_s) - \mathrm{log}_2(4)}\right \rfloor$, where $i_s$ is the input image dimension. For example, for input images of size $224\times224$, $i_s=224$ and $M=5$. Note that this number of layers is automatically given since our network is based on multiresolution analysis. The effect of choosing different levels, than the ones given by $M$ is analyzed during the hyperparameter study (Section~\ref{section:hyper}).
%
%
\paragraph*{Initial convolutional layers.}

As in every classification task, the proposed approach needs a discriminative representation of the data before the classification takes place. To obtain a discriminative feature set before the first downsampling of the signal, the architecture starts by extracting descriptors with two sequences of Conv-BN-ReLU, where Conv and BN stand for Convolution and Batch Normalization respectively, with kernel size $3 \times 3$ and with the same depth. The depth in these initial convolutional layers is one of the few hyper-parameters of DAWN. By fixing the depth and determining the number of decomposition levels, one can automatically obtain the depth of features maps of the last 2D lifting scheme for a given input image size.
%
\paragraph*{Loss function and constraints.}
%
%
End-to-end training is performed using the cross-entropy loss function,
in combination with some regularization terms to enforce a wavelet decomposition structure during training. The loss function takes the form of Eqn.~\ref{Equation:loss}, where $P$ denotes the number of classes, $y_i$  and $p_{i}$ are the binary ground-truth and the predicted probability for belonging to class $i$,  respectively. The regularization parameters $\lambda_1$ and $\lambda_2$ tune the strength of the regularization terms.
Also, $m_l^I$ and $m_l^C$  denote the mean of the input signal to the lifting scheme at level $l$ and the mean of the approximation sub-band at level $l$, respectively. And, $\mathbb{D}_l$ is the concatenation of the vectorized detail sub-bands at level $l$.
\begin{equation}\label{Equation:loss}
\begin{split}
    \mathbf{Loss} &=-\sum_{i=1}^{P}y_{i}\text{log}(p_{i}) \\
    &+ \lambda_1 \sum_{l=1}^M \mathrm{H}(\mathbb{D}_l) + \lambda_2\sum_{l=1}^M \|m_l^I-m_l^C\|_2^2.
\end{split}
\end{equation}
To promote low-magnitude detail coefficients~\cite{Litfting_loss1}, the first regularization term in Eqn.~\ref{Equation:loss} minimizes the sum of the Huber norm of $\mathbb{D}_l$ across all the decomposition levels. The choice of a Huber norm compared to $\ell_1$ is motivated by training stability. The second regularization term minimizes the sum of the $\ell_2$ norm of the difference between $m_l^I$ and $m_l^C$ across all the decomposition levels in order to preserve the mean of the input signal to form a proper wavelet decomposition~\cite{Litfting_loss1}.
\section{Experiments and Results}
 The evaluation of the DAWN model was analyzed on one texture dataset, KTH-TIPS2-b and two benchmarks datasets for object recognition task, namely, CIFAR-10 and CIFAR-100.  
 The obtained results are compared against different models commonly used for classification: ResNet~\cite{ResNet}; DenseNet~\cite{DenseNet} with growing factor of 12; a variant of VGG~\cite{VGG}, which adds batch normalization, global average pooling, and dropout.
The proposed architecture is also compared with previous networks using some multi-resolution analysis component: wavelet CNN (WCNN)~\cite{WCNN}, and Scattering network~\cite{Scattering}.
For this later one, we show the results of the handcrafted representation and the hybrid network that combines scattering transform on top of a Wide-Resnet. For KTH-TIPS2-b, T-CNN~\cite{T-CNN} results is shown as this architecture specifically tailored to texture analysis. 
 The training was done on multiple NVIDIA V100 Pascal GPUs with 12Gb of memory.
\subsection{Implementation}
An SGD optimizer with a momentum of $0.9$ is used for training. The initial learning rate is set to $0.03$ for all the databases. 
The batch size is set to $64$ and $16$ for the CIFAR databases and KTH-TIPS2-b, respectively. A learning rate decay of $0.1$ is applied on epochs $30$ and $60$ for KTH-TIPS2-b; and on epochs $150$ and $255$ for CIFAR. The number of epochs is set to $90$ and $300$ for KTH-TIPS2-b, and the CIFAR databases, respectively. The regularization parameters $\lambda_1$ and $\lambda_2$ are set to $0.1$ for all the experiments.
For the Scattering networks~\cite{Scattering} on the CIFAR databases, the original training setup has been used, as it achieves higher accuracy than the one obtained with the configuration proposed in this paper for the other architectures.
\subsection{Databases and Results }
\label{Section: Databases and Results}

\paragraph*{KTH-TIPS2-b} The KTH-TIPS Texture Database was developed by the Computational Vision and Active Perception Laboratory (CVAP) at the KTH Royal Institute of Technology in Stockholm \cite{KTH}. There are three versions of this dataset: KTH-TIPS, KTH-TIPS2-A and KTH-TIPS2-B. In this study, we work with the third version since it is the most widely used as benchmark in texture analysis. It contains 11 classes with four folders per class called samples, each sample has 108 images. As in other works \cite{WCNN}, one of the samples of each class was used for training and the rest sample folders were used for testing. The data augmentation consists in applying random cropping and mirroring operations. Table~\ref{tab:KTH_results} contains the average and standard deviation across different training sessions.

In this database, WCNN~\cite{WCNN} with 4 levels achieves better accuracy compared to T-CNN with a smaller number of trainable parameters. The proposed architecture with a depth of $16$  for the initial convolutional layers, achieves the same accuracy as WCNN but with a much smaller number of parameters. Note that the initial convolutional layers are essential for extracting meaningful feature representations, and without them the performance of the model drops significantly.

Scattering network with the handcrafted representation  (Scatter+FC) consist of using a scattering transform of spatial scale $5$ followed by a global average pooling and ending with a fully connected layer. This network configuration is very similar to the proposed network structure used for this database (Figure~\ref{fig:DAWN_architecture}). This network configuration achieves similar performance to the proposed approach with sightly less trainable parameters as the wavelets are not trainable. This result indicates that our architecture is able to learn representations that are similar to the scattering transform.

The proposed architecture performs better than DenseNet 13 and 22 BC with similar number of parameters. Note that for DenseNet, the number indicates the total number of layers used inside the network and BC meaning the use of the bottleneck compression approach~\cite{DenseNet}.
Scattering network with hybrid configuration (Scatter+WRN) increases significantly the number of trainable parameter compared to the handcrafted representation network. This hybrid configuration perform poorly as it overfit the dataset, and it has a highly dependence on the CNN architecture and the setup of hyperparameters.
\begin{table}[t]
    \caption{Comparison of accuracy results on the KTH-TIPS2-b database where all the network are trained from scratch without pre-trained information.}
    \vspace{5.0pt}
    \centering
    \begin{tabular}{|l|c|c|c|c|c|c|}
        \hline
         Architecture & \# param. & Avg. &Std.\\
         \hline
         T-CNN & 19'938'059 & 63.80 \% & 1.68 \\
         WCNN L4 & 10'211'811 & 68.83\% & 0.73\\
         \hline
         Scatter+WRN & 10'934'283 & 60.33 \% & 2.19 \\
         {Scatter+FC} & {22'484} & {68.57 \%} & {2.86} \\ 
         \hline
         DenseNet 22 BC & 74'684 & 65.71 \% & 1.35\\
         DenseNet 13 & 89'711 & 66.16 \% & 1.52\\
         \hline
         DAWN (no init.) & 2'894 & 58.60 \% & 4.10 \\
         DAWN (16 init.) & 71'227  & 68.88 \% &2.14 \\
         \hline
    \end{tabular}
    \label{tab:KTH_results}
\end{table}
\paragraph*{CIFAR}
CIFAR-10~\cite{Cifar} contains 60000 colour images of size $32\times 32$ belonging to 10 classes. The same partition used to train and test DenseNet~\cite{DenseNet} is used in this paper, \textit{i.e.} 50000 images for training and 10000 images for testing. CIFAR-100~\cite{Cifar} has 100 classes with 500 images per class. The data augmentation consists in applying random cropping with a padding of 4 pixels and horizontal mirroring operations.

Table~\ref{tab:CIFAR_Results} shows the best results of each architecture on these two databases. There are different DenseNet configurations available with a default growth value of 12. The configuration chosen for the comparison was the one with the closest number of parameters to that of the proposed model. The 18-layer ResNet architecture, after replacing the initial convolutional layers with a convolutional layer with stride $1$ and kernel size $3\times3$, is used for comparison. Those layers were removed because they are normally used to reduce the size of the image at the beginning of the network, which is not required for the small images of the CIFAR datasets. For WCNN, an experiment on varying the number of levels was conducted and the result of the best variant is reported in Table~\ref{tab:CIFAR_Results}.
{Scattering transform network configurations are the same used the original paper~\cite{Scattering} for these datasets.}

For the CIFAR databases, the proposed network uses three levels of lifting scheme, as the input image size is $32\times32$. Table~\ref{tab:CIFAR_Results} shows that increasing the number of initial convolutional filters tends to improve the accuracy performance. Therefore, it is up to the user to balance between a more compact network, in terms of number of parameters, and a network with better classification performance. DAWN architecture outperforms WCNN for both datasets even when the proposed architecture has significantly less number of parameters. {The scatter network with handcrafted representation (Scatter+MLP) achieves less accuracy than DAWN architecture as the wavelets are not learned.}

It also has a competitive accuracy for CIFAR-10 compared to VGG and ResNet architectures; furthermore, DAWN with a depth of $256$ for the initial convolution layers, outperforms the results in both architectures for CIFAR-100 dataset. {The scattering hybrid representation (Scatter+WRN) has a considerable higher number of parameters than the other architectures, and its performance is similar to VGG and ResNet for both datasets.}
In this application, the DenseNet architecture exhibits good performance due to its ability to retain relevant features through the entire network.
%
%
\begin{table}[t]
    \caption{Comparison of accuracy results on the CIFAR-10 and CIFAR-100 databases. {The number of trainable parameter are shown for CIFAR-100 database.}}
    \vspace{5.0pt}
    \centering
    \begin{tabular}{|l|c|c|c|}
        \hline
         Architecture & \# param. & CIFAR-10 & CIFAR-100\\ 
         \hline
         VGG (variation) & 15.0 M& 94.00 \% & 72.61 \% \\
         ResNet 18& 11.2 M& 94.25 \% & 73.30 \% \\
         DenseNet 40& 1.10 M& 94.73 \% & 75.25 \% \\
         DenseNet 100& 7.19 M& 95.90 \% & 79.8 \% \\
         \hline
         WCNN L3 & 2.28 M& 89.85 \%  & 65.17 \% \\
         {Scatter+WRN }&{45.5 M} &{92.31 \%} & {72.26 \%} \\
         {Scatter+MLP} & {17.0 M} & {81.90 \%} & {49.84 \%} \\
         \hline
         DAWN (16 init.) & 59.3 K & 86.04 \% & 56.7 \% \\
         DAWN (32 init.) &   0.21 M & 90.41 \% & 65.06 \% \\
         DAWN (64 init.) &  0.73 M & 92.69 \% & 70.57 \% \\
         DAWN (128 init.) & 2.79 M & \textbf{93.34} \% & 72.47 \% \\
         DAWN (256 init.) & 10.9 M & 92.02 \% & \textbf{74.04} \% \\
         \hline
    \end{tabular}
    \label{tab:CIFAR_Results}
\end{table}
\paragraph*{Hybrid network}
{As an additional experiment, the proposed multiresolution analysis can be combined with other network architecture. This hybrid network (DAWNN+WRN) consists in replacing the scattering transform by the 2D lifting schemes (Figure~\ref{fig:DAWN_architecture}) inside the Scatter+WRN architecture. This proposed hybrid architecture has similar number of trainable parameters than Scatter+WRN. On CIFAR databases, this architecture gets $93.76 \%$ and $74.88 \%$ of accuracy for CIFAR-10 and CIFAR-100, respectively, which is slightly higher compared to the one obtained by Scatter+WRN.}

\subsection{Hyperparameter Tuning}
\label{section:hyper}
DAWN network uses a few number of hyperparameters inside the architecture. Besides the initial convolution depth analyzed in Section \ref{Section: Databases and Results}, the other hyperparameters are the kernel size and the number of convolutional layers inside the updater and predictor of the lifting scheme. This section presents an analysis of the effect of these hyperparameters on the final architecture results. {For simplicity, the experiments are performed on CIFAR datasets using the DAWN architecture with 64 initial filters.}

\paragraph{Kernel size and number of convolutions} Both of these hyperparameters affect the lifting scheme module, whose role is to generate a mathematical function for the wavelet representation. The update operator $U$ needs to represent the frequency structure of the input signal, while the predictor $P$ needs to represent the spatial structure of the input signal. These hyperparameters also affect the final number of trainable parameters for the whole architecture. Table~\ref{tab:kernel_tuning} shows the effect when changing these hyperparameters: \textit{i)} the kernel size experiments were obtained with the U/P structure described in Figure~\ref{fig:lifting_wavelet_pooling} \textit{ii)} the number of hidden layers inside the module is generated by the repetition of the first convolutional layer of the U/P module. It is noticed that the performance results do not have a high variance for combinations of hyperparameters with similar number of trainable parameters.
\paragraph{Number of multiresolution analysis levels}
Table~\ref{tab:kernel_tuning} shows how the number of trainable parameters depends on the number of levels of the 2D adaptive lifting scheme. This table illustrates how the performance varies from not using any lifting scheme level (only initial convolutions), which results in poor performance, to using the maximum number of possible levels (according to Section~\ref{Section:DAWN}). As shown in Table~\ref{tab:kernel_tuning}, it is usually beneficial to use the maximum number of levels as it leads to higher accuracy values for both datasets. Note that in the CIFAR database, the input size is 32$\times$32, which makes makes the maximum number of possible levels equal to 3.

\begin{table}[t]
    \centering
    \caption{Results of tunning the DAWN architecture with 64 initial convolutions. The first table entry is the network configuration used to generate the results in Table~\ref{tab:CIFAR_Results}. The hyperparameters tested are kernel size (k), the number of hidden convolutional layers (h), and the number of levels (l). {The number of trainable parameter are shown for CIFAR-100 database.}}
    \vspace{5.0pt}
    \begin{tabular}{|l|c|c|c|}
    \hline
        Configuration & CIFAR-10 & CIFAR-100 & \# param.\\
    \hline
        (k=3, h=1, l=3) & 92.69 \% & 70.57 \% & 734'628 \\
    \hline
        (k=1, h=1, l=3) & 88.09 \% & 64.30 \%& 439'716\\
        (k=2, h=1, l=3) & 92.27 \% & 68.01 \%& 587'172\\
        (k=4, h=1, l=3) & 92.69 \% & 70.96 \%& 882'084\\
    \hline
        (k=3, h=2, l=3) & 92.58 \% & 70.51 \%& 918'564\\
        (k=3, h=3, l=3) & 92.46 \% & 68.85 \%& 1'140'900\\
        (k=3, h=4, l=3) & 92.35 \% & 68.19 \%& 1'363'236\\
    \hline
        (k=3, h=1, l=0) & 75.49 \%& 44.12 \%& 45'348 \\
        (k=3, h=1, l=1) & 90.53 \%& 66.71 \%& 275'108\\
        (k=3, h=1, l=2) & 92.17 \%& 70.42 \%& 504'868\\
    \hline
    \end{tabular}
    \label{tab:kernel_tuning}
\end{table}

%
%
\subsection{Visual Representation Results}
\begin{figure}[t]
    \centering
    \includegraphics[width=\linewidth]{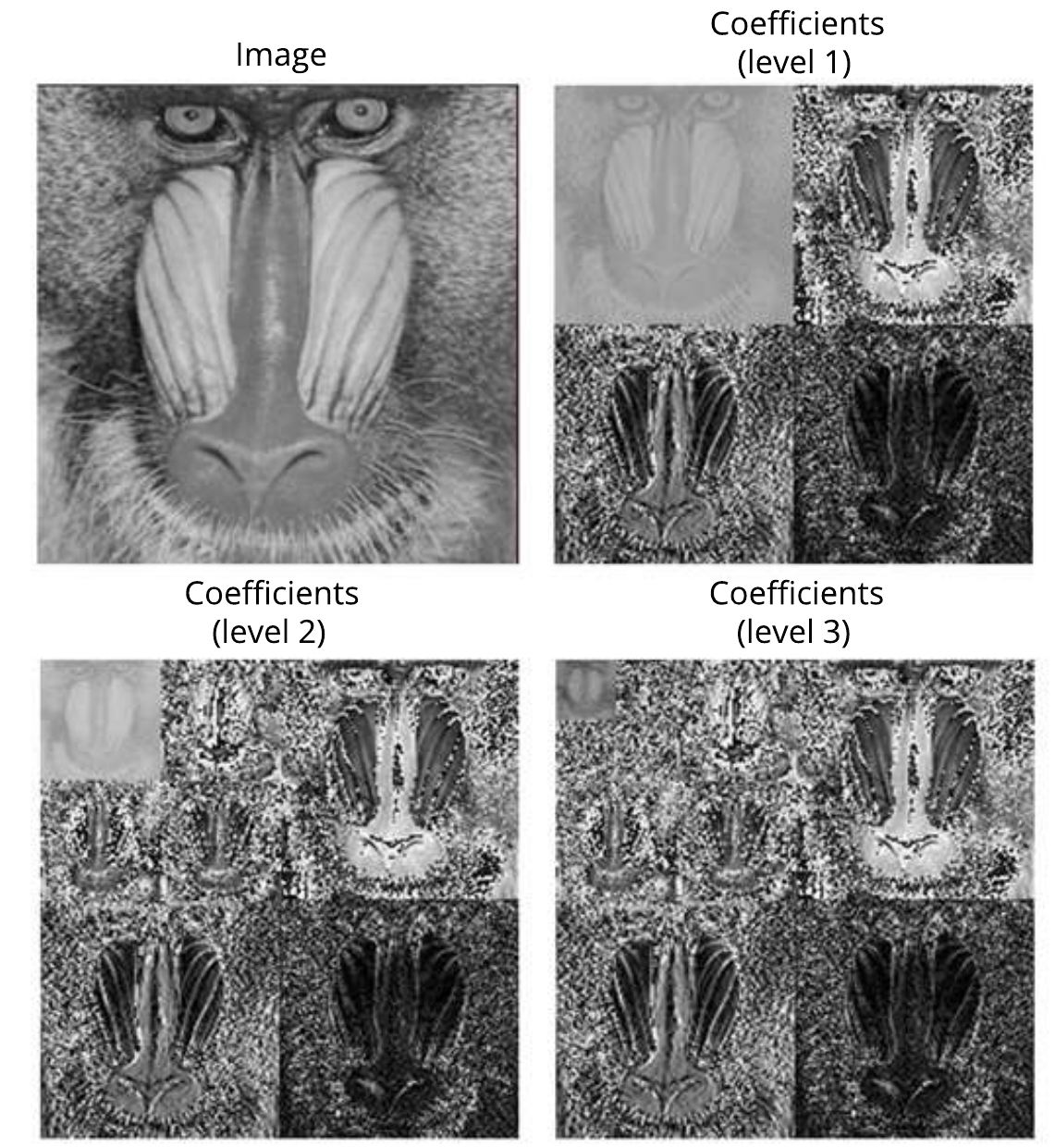}
    \caption{Results of extracting the coefficients for 3 decomposition levels of the 2D Adaptive Lifting Scheme in the DAWN architecture. The loss function applied is the same as in Eqn.~\ref{Equation:loss}. For visualization purposes, the LH, HL and HH sub-bands were multiplied by a factor of $10$.}
    \label{fig:dwnn_level0}
\end{figure}

The decomposition generated by the lifting scheme has a relevant visual representation as it is composed of approximation and details sub-bands of an input signal. Figure~\ref{fig:dwnn_level0} shows the visualization of the multiresolution analysis for different number of decomposition levels. To generate the visualizations presented in Figure \ref{fig:dwnn_level0}, the network was run without the initial convolutional layers on KTH database.

Many decomposition levels are very similar to traditional wavelet decomposition where the approximation sub-band captures the low-frequency information of the image while the detail sub-bands tend to capture high-frequency information. However, some sub-bands are slightly different as the loss function also minimize the cross-entropy loss function to ensure good classification performance (Section~\ref{Section:DAWN}).
\section{Discussion and Future work}
\paragraph*{Multiresolution analysis as a deep learning architecture}
{Analogous to DAWN architecture, Bruna and Mallat~\cite{Bruna2013} use a multiresolution analysis based on wavelet transform as a backbone of their architecture. Both, this work and our work, focus on the wavelet extraction as an operation invariant to deformation. In Bruna's work, the modulus is obtained from each wavelet coefficient at different levels. In DAWN architecture, the details coefficients per level of the wavelet transform are carried out to the end of the network.  One biggest difference between DAWN and the Scattering handcrafted representation is the ability of DAWN to learn the wavelet configuration. It is this ability that allows it to adapt to the data and perform equivalently across different datasets, as it was shown in Tables~\ref{tab:KTH_results} and ~\ref{tab:CIFAR_Results}.}
\paragraph*{Combining Multiresolution analysis with more traditional CNNs architectures} 
{The hybrid network with the proposed 2D lifting scheme shows the potential of improving the accuracy or reducing the number of trainable parameters for other networks. How to combining or incorporating more CNN features in the proposed network and keeping performance across the different datasets is an interesting work avenue.}

\paragraph*{Initial convolutions} At the moment, the architecture uses initial convolutional layers to increase the number of channels from the input image, which is a simple approach. Research using more advance architecture for this part of the proposed network is left as future work. Moreover, multiresolution analysis is usually apply on an image instead on a CNN output. Changing the order of the initial convolutions and the different lifting scheme might conduct to some exciting new architectures.

\section{Conclusions}
\label{SectionConclusions}
We presented the DAWN architecture, which combines the lifting scheme and CNNs to learn features using multiresolution analysis.
In contrast to the black-box nature of CNNs, the DAWN architecture is designed to extract a wavelet representation of the input at each decomposition level. Unlike traditional wavelets, the proposed model is data-driven so that it adapts to the input images. It is also trainable end-to-end and achieve state-of-the-art performance for texture classification with very limited number of trainable parameter. 
Interpreting convolution and pooling operations in CNNs as operations in multiresolution analysis helped us to systematically design a novel network architecture.
The performance of DAWN is comparable to that of state-of-the-art classification networks when tested on the CIFAR-10 and CIFAR-100 datasets. 
{\small
\bibliographystyle{ieee}
\bibliography{egbib}
}

\end{document}